\documentclass{article}

\usepackage{microtype}
\usepackage{graphicx}
\usepackage{booktabs} % for professional tables

\usepackage{hyperref}

\usepackage[accepted]{icml2025}

\usepackage{amsmath}
\usepackage{amssymb}
\usepackage{mathtools}
\usepackage{amsthm}

\usepackage[capitalize,noabbrev]{cleveref}

\theoremstyle{plain}

\theoremstyle{definition}

\theoremstyle{remark}

\usepackage[textsize=tiny]{todonotes}
\usepackage{multirow} 
\usepackage{caption}
\usepackage{subcaption}
\usepackage{pifont}

\icmltitlerunning{Submission and Formatting Instructions for ICML 2025}

\begin{document}

\twocolumn[
\icmltitle{DINT Transformer}



\icmlsetsymbol{equal}{*}

\begin{icmlauthorlist}
\icmlauthor{Yueyang Cang}{equal,tsinghua}
\icmlauthor{Yuhang Liu}{equal,tsinghua}
\icmlauthor{Xiaoteng Zhang}{tsinghua}
\icmlauthor{Erlu Zhao}{peking}
\icmlauthor{Shi Li}{tsinghua}
\end{icmlauthorlist}

\icmlaffiliation{tsinghua}{Tsinghua University, Beijing, China}
\icmlaffiliation{peking}{Peking University School of Public Health, Beijing, China}

\icmlcorrespondingauthor{Shi Li}{shilits@tsinghua.edu.cn}

\icmlkeywords{Natural Language Processing, Transformers, ICML}

\vskip 0.3in
]

\printAffiliationsAndNotice{\icmlEqualContribution}

\begin{abstract}
DIFF Transformer addresses the issue of irrelevant context interference by introducing a differential attention mechanism that enhances the robustness of local attention. However, it has two critical limitations: the lack of global context modeling, which is essential for identifying globally significant tokens, and numerical instability due to the absence of strict row normalization in the attention matrix. To overcome these challenges, we propose DINT Transformer, which extends DIFF Transformer by incorporating a differential-integral mechanism. By computing global importance scores and integrating them into the attention matrix, DINT Transformer improves its ability to capture global dependencies. Moreover, the unified parameter design enforces row-normalized attention matrices, improving numerical stability. Experimental results demonstrate that DINT Transformer excels in accuracy and robustness across various practical applications, such as long-context language modeling and key information retrieval. These results position DINT Transformer as a highly effective and promising architecture.
\end{abstract}

\begin{figure*}[htbp]
    \centering
    \includegraphics[width=1.0\textwidth]{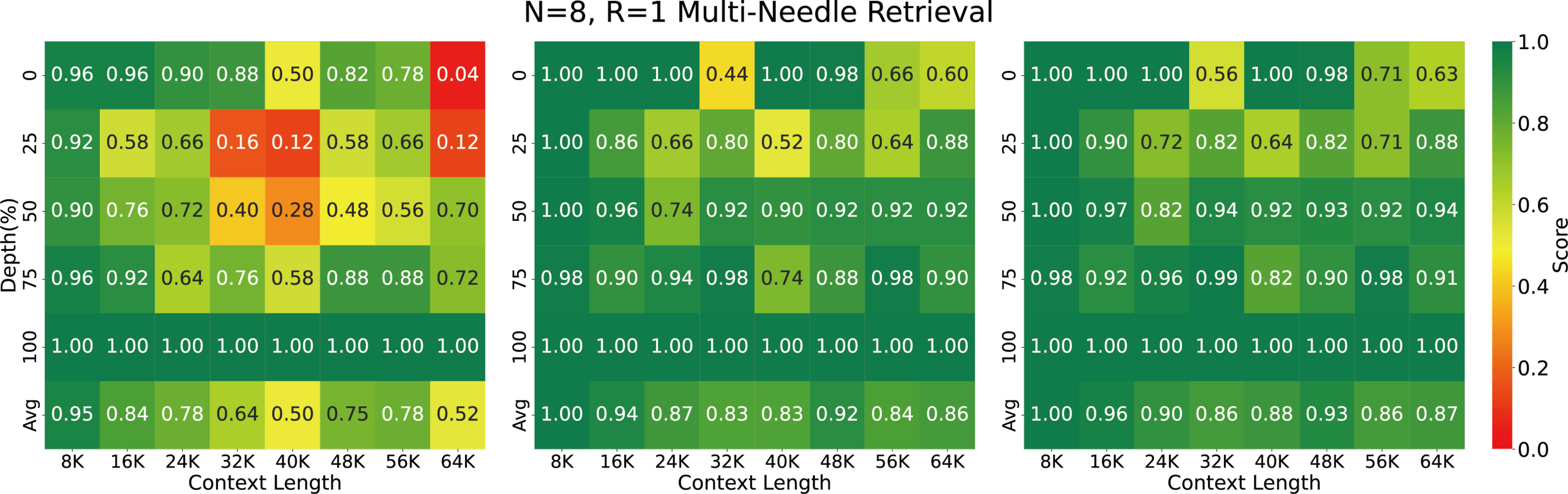}  % 请替换为图像的路径
    \caption{Transformer often over-attends to irrelevant context (i.e., attention noise). DINT Transformer not only eliminates noise but also strengthens the attention to globally important tokens, such as 'Newton' in the sentence.}
    \label{fig:diff_attention}
\end{figure*}

\section{Introduction}

Transformer\cite{vaswani2017attention}, as one of the most popular models in the field of artificial intelligence today, is widely used in natural language processing, computer vision, and other fields, especially with the application of decoder-only architectures in large language models (LLMs). Its core lies in the attention mechanism based on softmax, which assigns importance to different tokens in a sequence. However, recent research\cite{lu2021fantastically} has found that LLMs face the challenge of attention noise when accurately focusing on key information in the context.

To address the issue of attention noise, DIFF Transformer\cite{ye2024differential} introduces a differential attention mechanism that effectively suppresses the impact of irrelevant context by computing DIFFerence between two independent attention distributions. However, DIFF Transformer still has a significant limitation: DIFFerential operation causes the resulting attention matrix to fail in guaranteeing that the sum of each row equals one. This introduces numerical instability into the model's internal computations and may adversely affect the performance of downstream tasks.

In our study, we observe that many tokens within a sequence often rely on a few globally critical tokens for their semantic interpretation. For example, in a sentence, key elements such as the subject or main predicate verb often serve as semantic anchors, playing a crucial role in shaping the overall meaning of the sentence (as shown in Figure \ref{fig:diff_attention}). Based on this observation, we propose DINT Transformer, which extends DIFF Transformer by introducing an integral mechanism. This integral component computes global importance scores, enhancing the model's focus on critical tokens. Our proposed DINT Transformer not only reduces attention noise further by strengthening the focus on globally important tokens but also ensures row-normalized attention matrices through the parameter setup, resolving the numerical instability issue present in DIFF Transformer, thereby significantly improving model performance.

We conducted extensive experiments on tasks such as long-context language modeling and key information retrieval to evaluate the effectiveness of DINT Transformer. The results demonstrate that DINT Transformer consistently outperforms both the Transformer and DIFF Transformer, especially in long-sequence tasks, where its integral mechanism effectively captures global dependencies and further reduces attention noise. By ensuring row-normalized attention distributions, DINT Transformer provides an interpretable and robust attention mechanism, addressing key limitations of prior approaches. Furthermore, DINT Transformer enhances performance in downstream tasks like key information retrieval while maintaining scalability. These findings establish DINT Transformer as a powerful and efficient foundation for future advancements in sequence modeling and large language models.

\section{DINT Transformer}

\begin{figure*}[htp]
    \centering
    \includegraphics[width=0.9\textwidth]{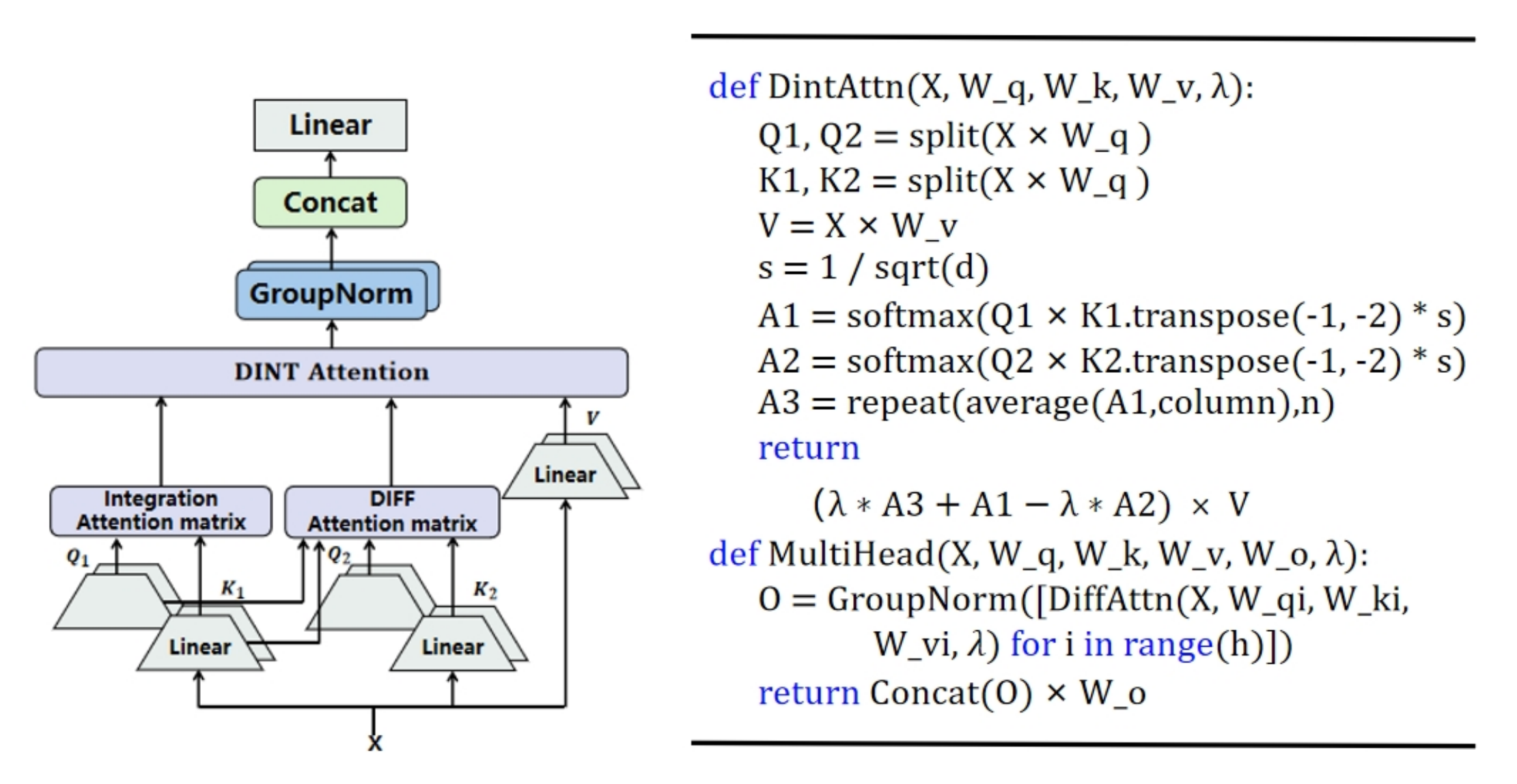} 
    \caption{Multi-head DINT Attention. DIFF Attention matrix implements reducing attention noise, while the Integration Attention matrix enhances global attention.}
    \label{fig:model}
\end{figure*}

DINT Transformer is designed as a robust architecture for sequence modeling, particularly for large language models (LLMs). The model consists of $L$ stacked layers, where each layer applies a DINT attention module followed by a feedforward network. Starting from token embeddings $X_0 \in \mathbb{R}^{N \times d_{\text{model}}}$, the input is progressively transformed through $L$ layers to produce the final output $X_L$. The key innovation lies in the addition of an integral mechanism within the attention module, which enables effective modeling of global dependencies while preserving numerical stability. The overall structure aligns with common practices, incorporating pre-RMSNorm\cite{zhang2019root} and SwiGLU\cite{ramachandran2017swish,shazeer2020glu} for enhanced performance following LLaMA\cite{touvron2023llama}. A diagram of the model architecture is shown in Figure ~\ref{fig:model}.

\begin{figure*}[h]
    \centering
    \begin{minipage}[b]{0.45\textwidth}
        \centering
        \includegraphics[width=\textwidth]{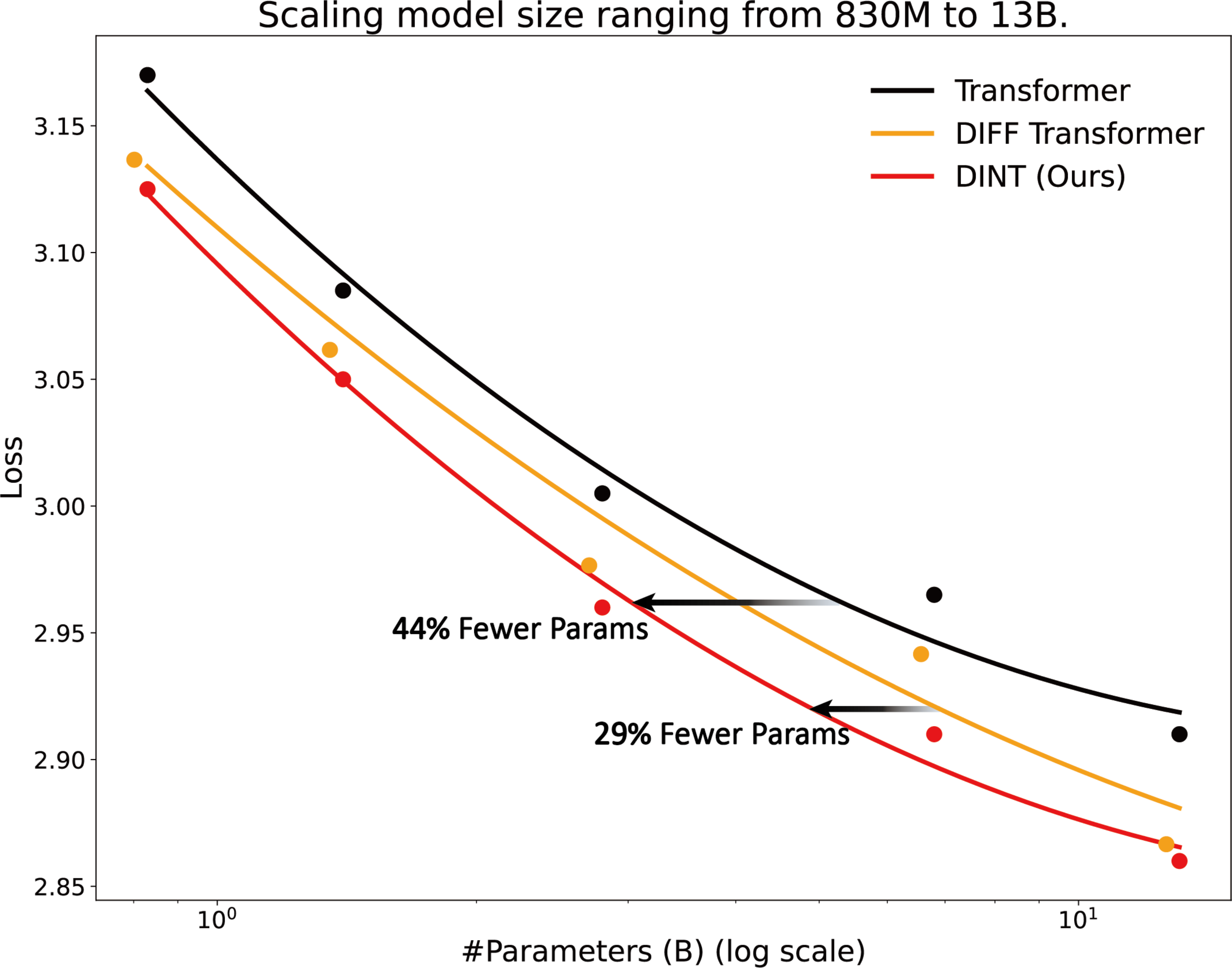}
        \caption*{(a) Scaling model size ranging from 830M to 13B.}
    \end{minipage}
    \hfill
    \begin{minipage}[b]{0.45\textwidth}
        \centering
        \includegraphics[width=\textwidth]{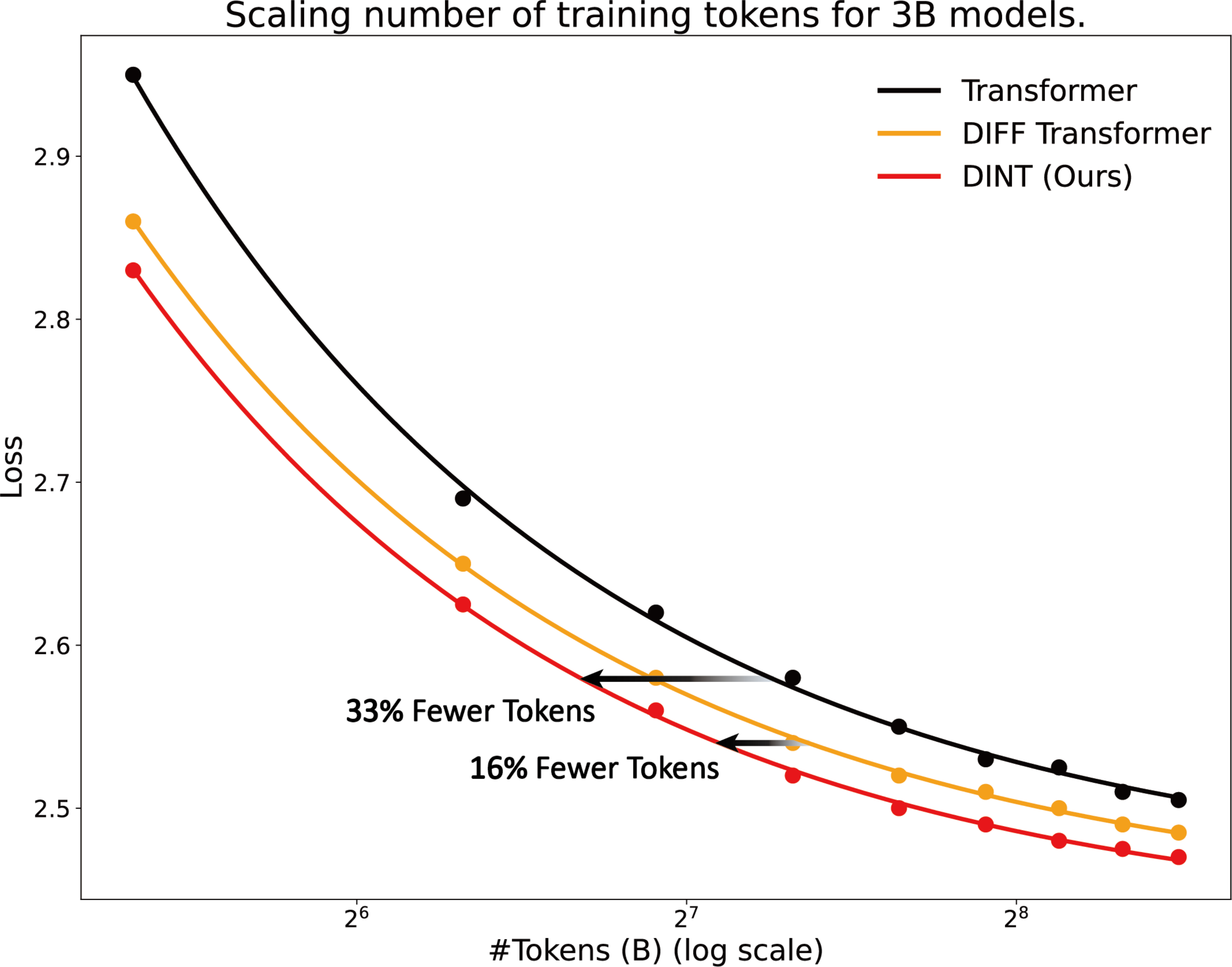}
        \caption*{(b) Scaling number of training tokens for 3B models.}
    \end{minipage}
    \caption{Language modeling loss of scaling up parameter count and training tokens. DINT Transformer outperforms other models, demonstrating that it requires fewer parameters or tokens to achieve comparable performance. (a) DINT Transformer matches the performance of larger models with fewer parameters. (b) DINT Transformer achieves comparable performance using significantly fewer training tokens.}
    \label{fig:scaling_comparison}
\end{figure*}
\subsection{DIFF Attention}

DIFF attention introduces a differential attention mechanism that reduces attention noise by leveraging the difference between two attention distributions. Specifically, given the input $X \in \mathbb{R}^{N \times d_{\text{model}}}$, it is projected to query, key, and value matrices:
\begin{equation}
    [Q_1; Q_2] = XW_Q, \quad [K_1; K_2] = XW_K, \quad V = XW_V,
\end{equation}
where $Q_1, Q_2, K_1, K_2 \in \mathbb{R}^{N \times d}$ and $V \in \mathbb{R}^{N \times 2d}$ are the projected matrices, and $W_Q, W_K, W_V \in \mathbb{R}^{d_{\text{model}} \times 2d}$ are learnable parameters. The differential attention operator computes the output as:
\begin{equation}
    \footnotesize
    \text{DiffAttn}(X) = \left( 
    \text{softmax}\left(\frac{Q_1 K_1^\top}{\sqrt{d}}\right) 
    - \lambda \cdot \text{softmax}\left(\frac{Q_2 K_2^\top}{\sqrt{d}}\right) 
    \right) V
\end{equation}

where $\lambda$ is a learnable scalar parameter. This differential mechanism effectively suppresses irrelevant context, enhancing the robustness of the attention scores by canceling common-mode noise, analogous to the operation of differential amplifiers in electrical engineering. To synchronize learning dynamics, $\lambda$ is re-parameterized as:
\begin{equation}
    \lambda = \exp(\lambda_{q1} \cdot \lambda_{k1}) - \exp(\lambda_{q2} \cdot \lambda_{k2}) + \lambda_{\text{init}},
\end{equation}
where $\lambda_{q1}, \lambda_{k1}, \lambda_{q2}, \lambda_{k2} \in \mathbb{R}^d$ are learnable vectors, and $\lambda_{\text{init}} \in (0, 1)$ is a constant used for initialization. Empirical results show that setting $\lambda_{\text{init}} = 0.8 - 0.6 \times \exp(-0.3 \cdot (l - 1))$, where $l \in [1, L]$ represents the layer index, works well in practice.

\subsection{DINT Attention}

DINT attention extends DIFF attention by introducing an integral mechanism, enhancing the model's ability to capture globally important information while maintaining numerical stability through row normalization in the final attention matrix. The signal attention matrix $A_{\text{1}}$ is computed using $Q_1$ and $K_1$:
\begin{equation}
    A_{\text{1}} = \text{softmax}\left(\frac{Q_1 K_1^\top}{\sqrt{d}}\right).
\end{equation}

The integral component computes global importance scores by averaging the signal attention weights column-wise:
\begin{equation}
    G = \frac{1}{N} \sum_{m=1}^N A_{\text{1}}[m, n],
\end{equation}
where $G \in \mathbb{R}^{1 \times N}$ is then expanded to match the dimensions of differential component:
\begin{equation}
    G_{\text{expanded}} = \text{repeat}(G, N),
\end{equation}
where $G_{\text{expanded}} \in \mathbb{R}^{N \times N}$ is obtained by repeating $G$ across $N$ rows.

DINT attention operator computes the output as:
\begin{equation}
    \text{DINTAttn}(X) = \left( A_{\text{diff}} + \gamma \cdot G_{\text{expanded}} \right) V,
\end{equation}
where $\gamma$ is a learnable scalar following DIFF Transformer, $A_{\text{diff}}$ is DIFF attention component, and $G_{\text{expanded}}$ is the expanded global importance scores matrix.

\textbf{Unified Parameter Setting.} By setting $\lambda$ and $\gamma$ to the same value, we ensure that the final attention matrix $A_{\text{final}}$ has rows that sum to 1. This row normalization guarantees numerical stability and consistency across the model, thusintaining data stability throughout the layers. This unified setting follows the parameterization method used in DIFF Transformer, further enhancing training stability.

\subsection{Multi-Head Differential Attention}

We also use the multi-head mechanism in DINT Transformer. Let $h$ denote the number of attention heads. We use different projection matrices $W_Q^i$, $W_K^i$, $W_V^i$, $i \in [1, h]$ for the heads. The scalar $\lambda$ is shared between heads within the same layer. Then the head outputs are normalized and projected to the final results as follows:
\begin{equation}
    \text{head}_i = \text{DiffAttn}(X; W_Q^i, W_K^i, W_V^i, \lambda)
\end{equation}
\begin{equation}
    \overline{\text{head}}_i = \text{LN}(\text{head}_i)
\end{equation}
\begin{equation}
    \text{MultiHead}(X) = \text{Concat}(\overline{\text{head}}_1, \cdots, \overline{\text{head}}_h) W_O
\end{equation}
where $W_O \in \mathbb{R}^{d_{\text{model}} \times d_{\text{model}}}$ is a learnable projection matrix, $\text{LN}(\cdot)$ uses RMSNorm for each head, and $\text{Concat}(\cdot)$ concatenates the heads together along the channel dimension. Unlike DIFF Transformer, we do not apply an additional multiplier to the outputs of each head, as the unified parameter setting in DINT Transformer already ensures numerical stability and consistency. The number of heads is set as $h = d_{\text{model}} / 2d$, where $d$ is the head dimension of the Transformer, to ensure that the parameter count and computational complexity are aligned.

\textbf{Headwise Normalization.} Figure~\ref{fig:model} illustrates the use of GroupNorm\cite{wu2018group} within the attention mechanism to stabilize training. Although Layer Normalization (LN) is applied independently to each attention head, the sparse nature of differential attention often leads to varied statistical patterns across heads. By normalizing each head individually before the concatenation step, LN ensures more consistent gradient statistics, which contributes to improved training stability\cite{qin2022devil,wang2023magneto}.

\subsection{Overall Architecture}

The overall architecture stacks $L$ layers, where each layer contains a multihead differential attention module and a feedforward network module. We describe DINT Transformer layer as:
\begin{equation}
    Y^l = \text{MultiHead}(\text{LN}(X^l)) + X^l
\end{equation}
\begin{equation}
    X^{l+1} = \text{SwiGLU}(\text{LN}(Y^l)) + Y^l
\end{equation}
where $\text{LN}(\cdot)$ is RMSNorm, and $\text{SwiGLU}(X)$ is defined as:
\[
\text{SwiGLU}(X) = (\text{swish}(XW_G) \odot XW_1)W_2,
\]
where $W_G, W_1 \in \mathbb{R}^{d_{\text{model}} \times \frac{8}{3} d_{\text{model}}}$, and $W_2 \in \mathbb{R}^{\frac{8}{3} d_{\text{model}} \times d_{\text{model}}}$ are learnable matrices.

\section{Experiments}

In this study, we evaluate DINT Transformer through a series of experiments, comparing it with DIFF Transformer and other baseline models. Since DINT Transformer does not introduce new learnable parameters, only increasing computational complexity, its parameter count remains unchanged. Therefore, the model configurations used in the comparison were chosen to be the same as those of DIFF Transformer. Our experiments show that by enhancing attention to globally significant tokens, DINT Transformer effectively reduces attention noise. Additionally, DINT Transformer exhibits stronger stability compared to DIFF Transformer, leading to improved performance across tasks such as long-sequence modeling, key information retrieval, and in-context learning.

\subsection{Language Modeling Evaluation}

We trained a 3B DINT Transformer language model using the same configuration settings as the 3B DIFF Transformer language model. The model settings are shown in Table \ref{tab:config}.

\begin{table}[h]
    \centering
    \begin{tabular}{|l|c|}
    \hline
    \textbf{Params} & \textbf{Values} \\ \hline
    Layers          & 28              \\ 
    Hidden size     & 3072            \\ 
    FFN size        & 8192            \\ 
    Vocab size      & 100,288         \\ 
    Heads           & 12              \\ 
    Adam $\beta$    & (0.9, 0.95)     \\ 
    LR              & $3.2 \times 10^{-4}$ \\ 
    Batch size      & 4M              \\ 
    Warmup steps    & 1000            \\ 
    Weight decay    & 0.1             \\ 
    Dropout         & 0.0             \\ \hline
    \end{tabular}
    \caption{Configuration settings used for the 3B-size DINT Transformer and DIFF Transformer models.}
    \label{tab:config}
\end{table}

\textbf{Results.} Table ~\ref{tab:results} presents the zero-shot evaluation results on the LM Eval Harness benchmark \cite{gao10256836framework}. We compare DINT Transformer with other state-of-the-art Transformer-based models, including OpenLLaMA-v2-3B \cite{geng2023openllama}, StableLM-base-alpha-3B-v2 \cite{tow2023stablelm}, and StableLM-3B-4E1T \cite{stableLM}. All models were trained with 1T tokens under similar configurations to ensure a fair comparison. The results clearly highlight that DINT Transformer outperforms these models across various downstream tasks, demonstrating its exceptional ability to capture both local and global dependencies.

\begin{table*}[h]
    \centering
    \resizebox{\textwidth}{!}{%
    \begin{tabular}{|l|c|c|c|c|c|c|c|c|}
    \hline
    \textbf{Model} & \textbf{ARC-C} & \textbf{ARC-E} & \textbf{BoolQ} & \textbf{HellaSwag} & \textbf{OBQA} & \textbf{PIQA} & \textbf{WinoGrande} & \textbf{Avg} \\ \hline
    OpenLLaMA-3B-v2  & 33.9 & 67.6 & 65.7 & 70.0 & 26.6 & 76.7 & 62.9 & 57.5 \\ 
    StableLM-base-alpha-3B-v2  & 32.4 & 67.3 & 64.6 & 68.6 & 27.1 & 76.0 & 63.0 & 57.0 \\ 
    StableLM-3B-4E1T & -- & 66.6 & -- & -- & -- & 76.8 & 63.2 & -- \\ 
    DIFF-3B & 37.8 & 72.9 & 69.0 & 71.4 & 29.0 & 76.8 & 70.1 & 60.6 \\ 
    \textbf{DINT-3B} & \textbf{39.0} & \textbf{74.1} & \textbf{70.5} & \textbf{72.8} & \textbf{30.3} & \textbf{77.2} & \textbf{71.8} & \textbf{62.2} \\ \hline
    \end{tabular}%
    }
    \caption{Eval Harness accuracy compared with well-trained Transformer language models. The results indicate the superior performance of DINT Transformer over other models across a range of tasks.}
    \label{tab:results}
\end{table*}

\subsection{Scalability Compared with Transformer}

\begin{table}[h]
    \centering
    \begin{tabular}{|l|c|c|c|}
        \hline
        \textbf{Size} & \textbf{Hidden Dim.} & \textbf{\#Layers} & \textbf{\#Heads} \\ \hline
        830M & 1536 & 24 & 8 \\ 
        1.4B & 2048 & 24 & 8 \\ 
        2.8B & 2560 & 32 & 10 \\ 
        6.8B & 4096 & 32 & 16 \\ 
        13.1B & 5120 & 40 & 20 \\ \hline
    \end{tabular}
    \caption{Model configurations for different sizes, including hidden dimension, number of layers, and number of attention heads. Each model was trained with a sequence length of 2048 and a batch size of 0.25 million tokens, for a total of 40K training steps.}
    \label{tab:model_configs}
\end{table}

We evaluated the scalability of DINT Transformer compared to the standard Transformer and DIFF Transformer, specifically focusing on language modeling tasks. This evaluation involved scaling both model size and the number of training tokens. We adopted an enhanced Transformer architecture similar to LLaMA, ensuring a fair comparison by using identical experimental setups.

\textbf{Scaling Model Size} As shown in Figure \ref{fig:scaling_comparison}(a), DINT Transformer consistently outperformed both Transformer and DIFF Transformer across various model sizes (see Table \ref{tab:model_configs} for model configurations). Specifically, DINT Transformer achieved comparable validation loss to the Transformer with 44\% fewer parameters and matched the performance of DIFF Transformer with 29\% fewer parameters. This demonstrates the superior efficiency and scalability of DINT Transformer in terms of parameter usage.

\textbf{Scaling Training Tokens} Figure \ref{fig:scaling_comparison}(b) shows the results of scaling the number of training tokens. The fitted curves indicate that DINT Transformer achieved comparable performance to the Transformer with 33\% fewer training tokens. Additionally, DINT Transformer outperformed DIFF Transformer with 16\% fewer training tokens. These results highlight the significant data efficiency of DINT Transformer, achieving equivalent or superior results with considerably fewer resources.

\subsection{Key Information Retrieval}

The Needle-In-A-Haystack test \cite{kamradt2023needle} is used to evaluate the ability of models to extract key information from long contexts. Following the protocol of LWM \cite{liu2024world} and Gemini 1.5 \cite{reid2024gemini}, "needles" are short sentences that assign a unique number to a city. The objective is to retrieve these numbers based on a given query.

We position the answer needle at different depths within the context (0\%, 25\%, 50\%, 75\%, 100\%), while other needles are placed randomly. Each combination of depth and context length is evaluated over 50 samples, and the average accuracy is reported.

\begin{table}[h]
    \centering
    \resizebox{0.45\textwidth}{!}{%
    \begin{tabular}{|l|c|c|c|c|}
        \hline
        \multirow{2}{*}{\textbf{Model}} & \textbf{$N = 1$} & \textbf{$N = 2$} & \textbf{$N = 4$} & \textbf{$N = 6$} \\ 
                                      & \textbf{$R = 1$} & \textbf{$R = 2$} & \textbf{$R = 2$} & \textbf{$R = 2$} \\ \hline
        Transformer                   & \textbf{1.00}    & 0.85            & 0.62            & 0.55            \\ 
        DIFF                           & \textbf{1.00}    & 0.92            & 0.84            & 0.85            \\ 
        DINT                   & \textbf{1.00}    & \textbf{0.96}    & \textbf{0.89}    & \textbf{0.88}    \\ \hline
    \end{tabular}%
    }
    \caption{Multi-needle retrieval accuracy in 4K length contexts, averaged over the answer needle positions. $N$ represents the number of needles, and $R$ denotes the number of query cities.}
    \label{tab:multi_needle_retrieval}
\end{table}

\begin{figure*}[htbp]
    \centering
    \includegraphics[width=0.9\textwidth]{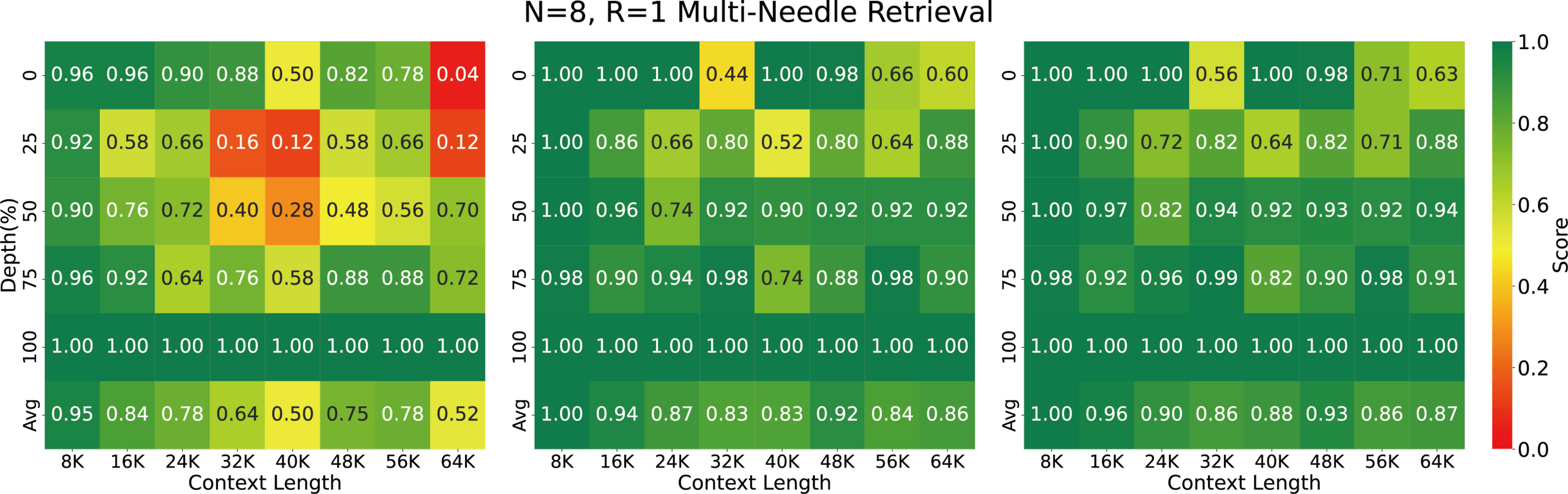}  % 请替换为图像的路径
    \caption{Multi-needle retrieval results in 64K length.}
    \label{fig:multi_needle_results}
\end{figure*}

\textbf{Retrieve from 4K Context Length} We evaluated the multi-needle retrieval task using 4K-length contexts, inserting $N = 1, 2, 4, 6$ needles and retrieving $R = 1, 2$ needles. The models used for evaluation were trained with an input length of 4K. As shown in Table \ref{tab:multi_needle_retrieval}, DINT Transformer consistently outperforms the other models. Particularly at $N = 6, R = 2$, DINT achieves an accuracy of $0.88$, significantly better than Transformer and DIFF models, indicating its superior ability to retrieve key information amidst distracting contexts.

\textbf{Retrieve from 64K Context Length} As shown in Figure \ref{fig:multi_needle_results}, the context lengths evaluated range from 8K to 64K, with the configuration set to $N = 8$, $R = 1$. We evaluated the 3B-scale model with extended context (as described in Section 3.3). The accuracy is reported across different answer needle depths (y-axis) and context lengths (x-axis). The bottom row shows the average accuracy across all depths. From the figure, it can be observed that DINT Transformer consistently performs well across varying context lengths and needle depths. Notably, at a 40K context length and 25\% needle depth, DINT Transformer shows a 52\% improvement in accuracy compared to Transformer and a 12\% improvement compared to DIFF Transformer.

\textbf{Attention Score Analysis} Table \ref{tab:attention_dint} presents the attention scores assigned to the answer span and the noise context in the key information retrieval task. These scores reflect the model's ability to focus on relevant information while ignoring irrelevant noise. We compare the normalized attention scores for different depths (i.e., positions) of the target answer within the context. The results show that DINT Transformer allocates significantly higher attention to the correct answer span and exhibits a substantial reduction in attention noise.

\begin{table*}[h]
\centering
\begin{tabular}{lcccccccccc}
\toprule
\textbf{Model} & \multicolumn{5}{c}{\textbf{Attention to Answer↑}} & \multicolumn{5}{c}{\textbf{Attention Noise↓}} \\
 & 0\% & 25\% & 50\% & 75\% & 100\% & 0\% & 25\% & 50\% & 75\% & 100\% \\
\midrule
Transformer & 0.03 & 0.03 & 0.03 & 0.07 & 0.09 & 0.51 & 0.54 & 0.52 & 0.49 & 0.49 \\
DIFF & 0.27 & 0.30 & 0.31 & 0.32 & 0.40 & 0.01 & 0.02 & 0.02 & 0.02 & 0.01 \\
DINT (Ours) & \textbf{0.35} & \textbf{0.38} & \textbf{0.40} & \textbf{0.41} & \textbf{0.45} & \textbf{0.01} & \textbf{0.01} & \textbf{0.01} & \textbf{0.01} & \textbf{0.01} \\
\bottomrule
\end{tabular}
\caption{Attention scores allocated to answer spans and noise context in the key information retrieval task. The target answer is inserted at varying depths within the context. DINT Transformer allocates more attention to relevant information and effectively minimizes attention noise.}
\label{tab:attention_dint}
\end{table*}

\subsection{In-Context Learning}

We investigate in-context learning from two main angles: the performance on many-shot classification tasks and the model’s ability to maintain robustness when utilizing context. In-context learning is an essential trait of language models, reflecting their capability to make effective use of the provided input context.

\begin{figure*}[htbp]
    \centering
    \begin{minipage}{0.4\textwidth}
        \centering
        \includegraphics[width=\linewidth]{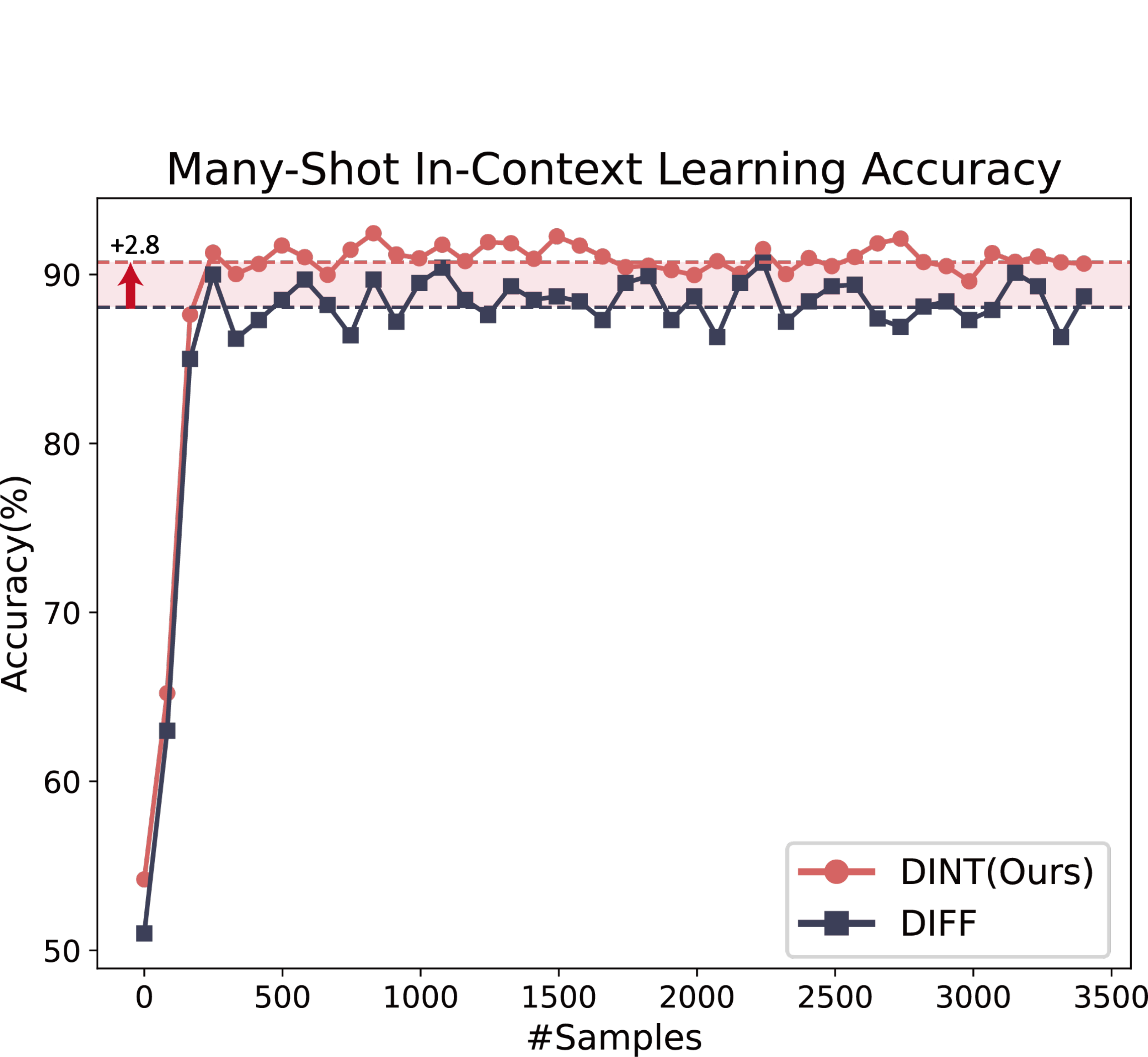} % Replace with your image file
        \subcaption{TREC dataset (6 classes)} \label{fig:trec}
    \end{minipage}%
    \hspace{0.5cm}
    \begin{minipage}{0.4\textwidth}
        \centering
        \includegraphics[width=\linewidth]{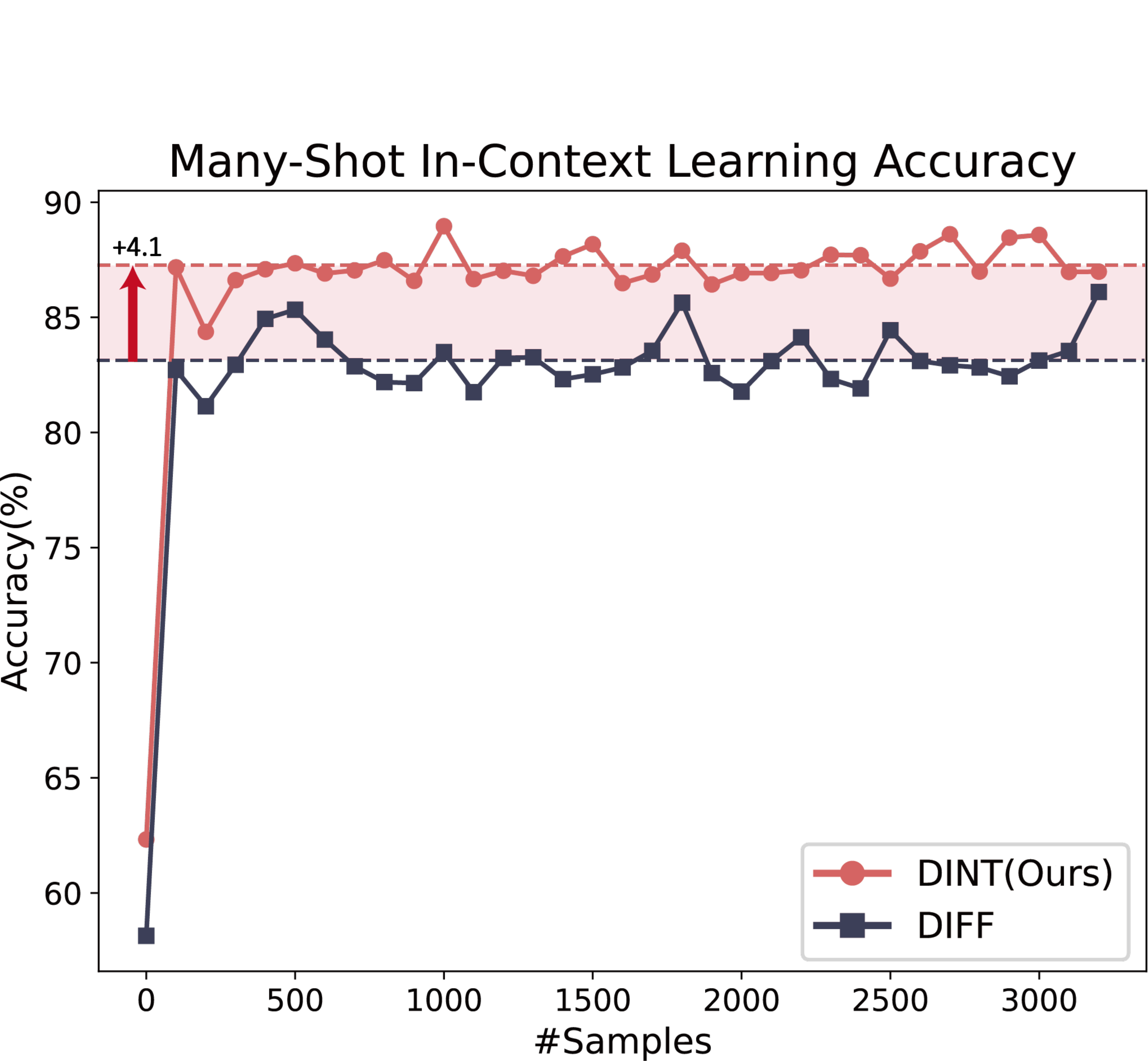} % Replace with your image file
        \subcaption{TREC-fine dataset (50 classes)} \label{fig:trec_fine}
    \end{minipage}
    \vskip\baselineskip
    \begin{minipage}{0.4\textwidth}
        \centering
        \includegraphics[width=\linewidth]{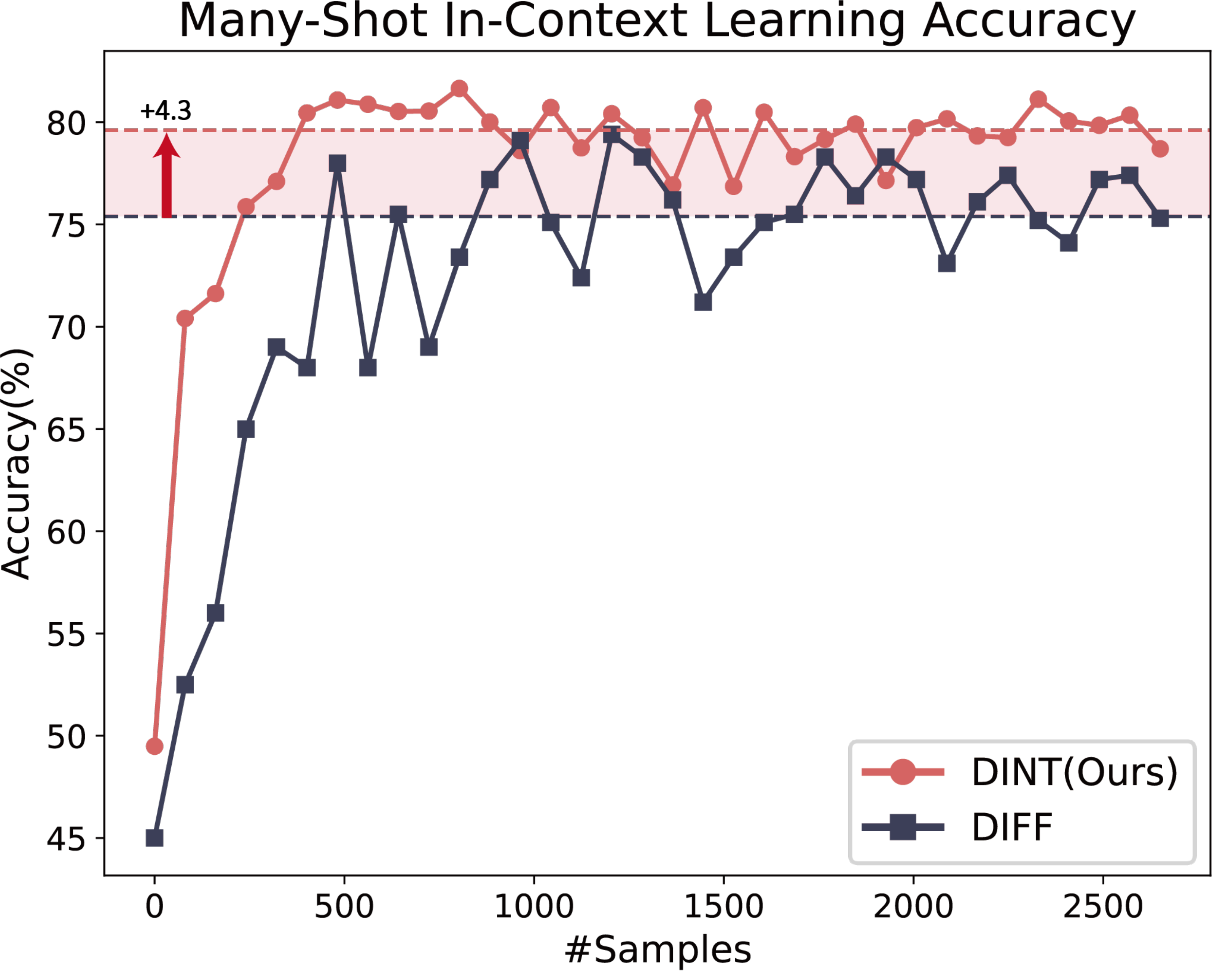} % Replace with your image file
        \subcaption{Banking-77 dataset (77 classes)} \label{fig:banking_77}
    \end{minipage}%
    \hspace{0.5cm}
    \begin{minipage}{0.4\textwidth}
        \centering
        \includegraphics[width=\linewidth]{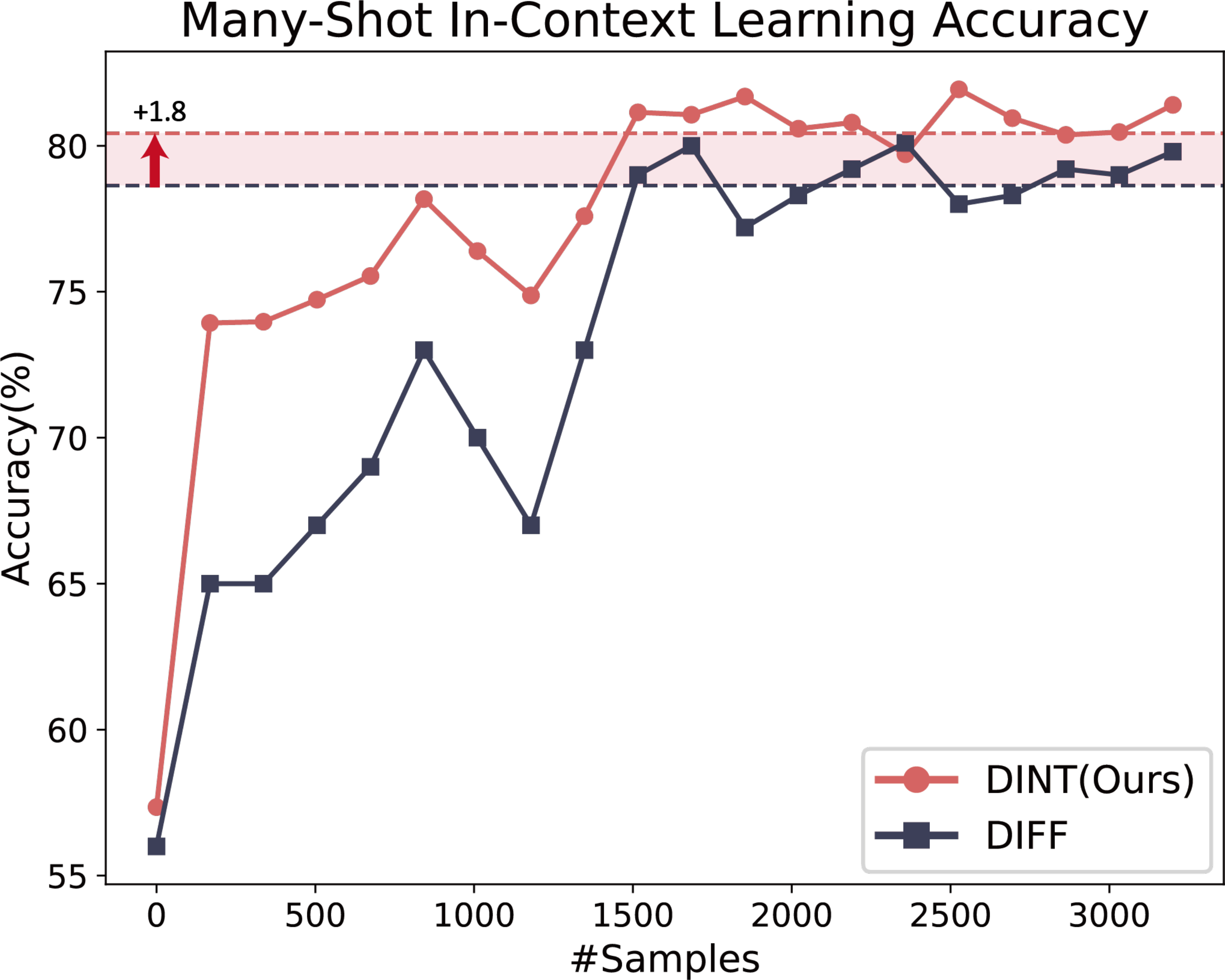} % Replace with your image file
        \subcaption{Clinic-150 dataset (150 classes)} \label{fig:clinic_150}
    \end{minipage}
    \caption{Accuracy of many-shot in-context learning across four datasets, with demonstration examples increasing from 1-shot up to a total of 64K tokens. The dashed lines indicate the average accuracy once the model's performance stabilizes.} 
    \label{fig:many_shot_incontext_learning}
\end{figure*}

\begin{figure*}[h!]
    \centering
    % 插入第一张图
    \begin{minipage}[b]{0.48\textwidth}
        \centering
        \includegraphics[width=\textwidth]{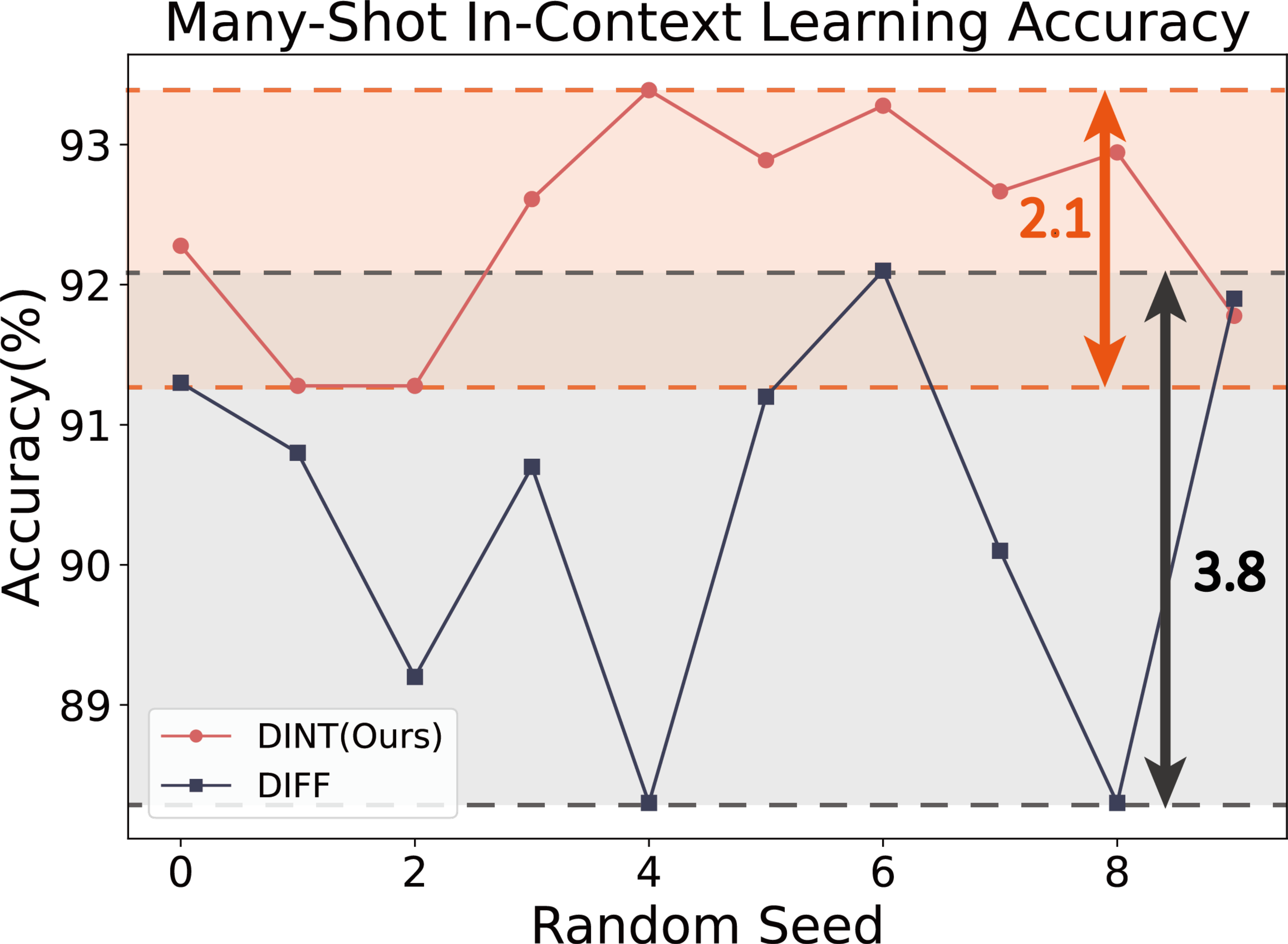}  
        \caption*{(a) Examples are randomly arranged.}
    \end{minipage}
    \hfill
    % 插入第二张图
    \begin{minipage}[b]{0.48\textwidth}
        \centering
        \includegraphics[width=\textwidth]{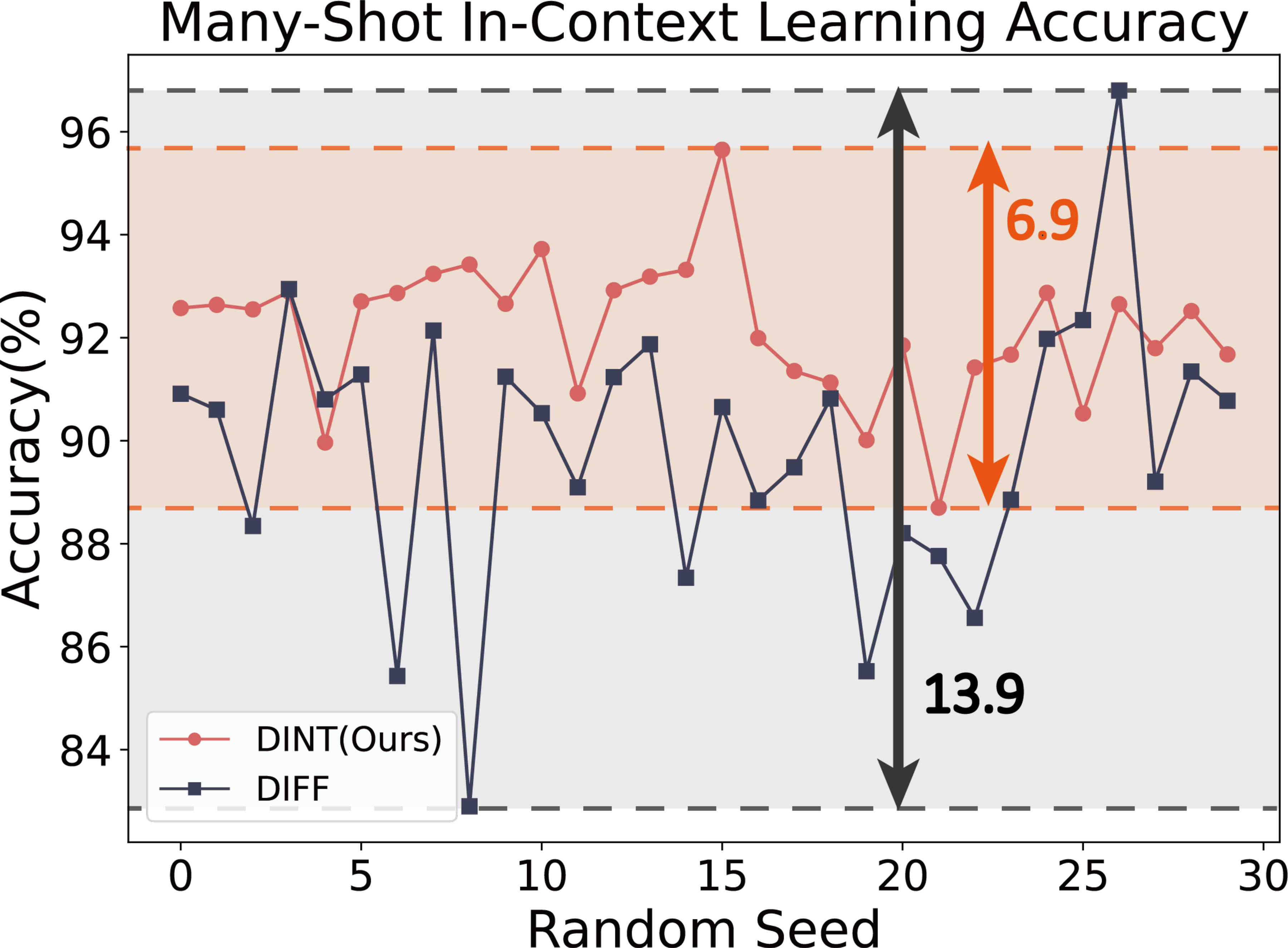}
        \caption*{(b) Examples are arranged alternately by class.}
    \end{minipage}
    \caption{Many-shot in-context learning accuracy on four datasets. The accuracy for both DIFF Transformer and DINT (Ours) models is presented, showing performance improvements across different numbers of demonstration samples.}
    \label{fig:Robustness}
\end{figure*}

\begin{table*}[h]
\centering
\begin{tabular}{lccccccc}
\toprule
\multirow{2}{*}{\textbf{Model}} & \multirow{2}{*}{\textbf{\#Heads}} & \multirow{2}{*}{\textbf{d}} & \multirow{2}{*}{\textbf{GN}} & \multirow{2}{*}{\textbf{Valid. Set↓}} & \multicolumn{2}{c}{\textbf{Fine-Grained Slices}} \\ 
 &  &  &  &  & \textbf{AR-Hit↓} & \textbf{Others↓} \\
\midrule
DIFF  & 8  & 128 & \ding{51} & 3.062 & 0.880 & 3.247 \\
\textminus GroupNorm  & 8  & 128 & \ding{55} & 3.122 & 0.911 & 3.309 \\
with $\lambda_{\text{init}} = 0.8$  & 8  & 128 & \ding{51} & 3.065 & 0.883 & 3.250 \\
with $\lambda_{\text{init}} = 0.5$  & 8  & 128 & \ding{51} & 3.066 & 0.882 & 3.251 \\
\midrule
\textbf{DINT (Ours)} & 8  & 128 & \ding{51} & \textbf{3.055} & \textbf{0.875} & \textbf{3.243} \\
\textminus GroupNorm & 8  & 128 & \ding{55} & 3.075 & 0.893 & 3.256 \\
with $\lambda_{\text{init}} = 0.8$  & 8  & 128 & \ding{51} & 3.056 & 0.877 & 3.245 \\
with $\lambda_{\text{init}} = 0.5$ & 8  & 128 & \ding{51} & 3.058 & 0.878 & 3.245 \\
\bottomrule
\end{tabular}
\caption{Ablation Studies of 1.4B-Size Models.}
\label{tab:ablation_dint_diff}
\end{table*}

\textbf{Many-Shot In-Context Learning} As presented in Figure~\ref{fig:many_shot_incontext_learning}, we compare the accuracy of many-shot classification between DIFF Transformer and our DINT Transformer architecture. We evaluate the 3B-size language models that support 64K input length. We follow the evaluation protocol of \cite{bertsch2024in} and use constrained decoding \cite{ratner2023parallel}. The number of demonstration samples is incrementally increased from 1-shot until the total length reaches 64K tokens. Specifically, we evaluate the models on the following datasets: TREC \cite{hovy2001toward} with 50 classes, Banking-77 \cite{casanueva2020efficient} with 77 classes, and Clinic-150 \cite{larson2019evaluation} with 150 classes. The results show that DINT Transformer consistently outperforms DIFF Transformer across all datasets and varying numbers of demonstration samples. The improvement in average accuracy is substantial, with DINT achieving 2.8\% higher accuracy on TREC, 4.1\% on TREC-Fine, 4.3\% on Banking-77, and 1.8\% on Clinic-150.

\textbf{Robustness of In-Context Learning} Figure ~\ref{fig:Robustness} presents a comparison of the robustness between DIFF Transformer and DINT Transformer in the context of in-context learning. By analyzing how performance varies with different order permutations of the same set of demonstration examples, we find that smaller performance fluctuations reflect greater robustness and a reduced risk of catastrophic degradation. The evaluation protocol remains consistent with the previously outlined methodology. Figure ~\ref{fig:Robustness} displays the results of this analysis on the TREC dataset. We examine two prompt configurations: randomly shuffled examples and examples arranged by class in an alternating pattern. In both configurations, DINT Transformer consistently shows smaller performance fluctuations compared to DIFF Transformer, demonstrating that our approach enhances robustness in in-context learning tasks.

\subsection{Ablation Studies}

We perform ablation studies using 1.4B-parameter language models, with the same training setup as the 1.4B model in Section 3.2. Both models have 24 layers, with 16 attention heads for Transformer and 8 for DIFF Transformer, each having a head dimension of 128.

Table~\ref{tab:ablation_dint_diff} reports the fine-grained loss on the validation set, breaking it into two components: "AR-Hit" and "Others." "AR-Hit" evaluates the model's ability to recall previously seen n-grams, while "Others" represents tokens that are either frequent or not recalled from the context.

As shown in Table \ref{tab:ablation_dint_diff}, we performed ablation studies on various design choices in DINT Transformer and compared them with Transformer variants. All models are of similar size and training FLOPs for a fair comparison. The results indicate that our method outperforms DIFF Transformer in both overall loss and fine-grained loss. When GroupNorm is removed, the performance of DIFF Transformer is significantly affected, while DINT Transformer shows a smaller impact. This is because we ensure the row normalization of the attention matrix, which improves the model's overall robustness. Additionally, when using constant initialization for lambda, we observe a slight decrease in performance, but the model still maintains a high level of performance. This demonstrates the effectiveness of our initialization method and shows that the model is robust to different initialization choices.

\section{Conclusions}

We propose DINT Transformer, which integrates global attention statistics into DIFF Transformer to reduce noise and enhance focus on key words. This improves the model's ability to capture global information, ensuring better stability and scalability. Experiments show DINT Transformer excels in long-sequence modeling, key information retrieval, and in-context learning, making it highly promising for NLP tasks requiring global context awareness.

\bibliography{example_paper}
\bibliographystyle{icml2025}
\end{document}